\documentclass[preprint,12pt]{elsarticle}
\usepackage[top=2cm, bottom=2cm, left=3cm, right=3cm]{geometry}

\usepackage{amssymb}
\usepackage{amsmath}
\usepackage{booktabs}
\usepackage{multirow}
\usepackage{algorithm}
\usepackage{algorithmic}
\usepackage{hyperref}
\usepackage[T1]{fontenc}
\usepackage{comment}

\newsavebox\CBox
\def\textBF#1{\sbox\CBox{#1}\resizebox{\wd\CBox}{\ht\CBox}{\textbf{#1}}}

\usepackage{colortbl}  
\usepackage{xcolor}
\definecolor{Gray}{gray}{0.9}

\journal{Knowledge-Based Systems}

\begin{document}

\begin{frontmatter}



\title{Learning Dynamic Representations via An Optimally-Weighted Maximum Mean Discrepancy Optimization Framework for Continual Learning}


\author[1]{Kaihui Huang} 
\author[2]{RunQing Wu}
\author[3]{JinHui Sheng}
\author[4]{HanYi Zhang}
\author[5]{Ling Ge}
\author[6]{JinYu Guo}

\author[7]{Fei Ye \texorpdfstring{\corref{cor1}}{}}
\ead{feiye@uestc.edu.cn}
\cortext[cor1]{Corresponding author}

\affiliation[1]{organization={School of Information and Software Engineering, University of Electronic Science and Technology of China}}
\affiliation[2]{organization={School of Mechanical Engineering, Huazhong University of Science and Technology}, country={China}}
\affiliation[3]{organization={Xihua University}, country={China}}
\affiliation[4]{organization={School of Computation, Information and Technology, Technische Universität München}, country={Germany}}
\affiliation[5]{organization={China Mobile Communications Group Chongqing Co., Ltd.}}
\affiliation[6]{organization={School of Information and Software Engineering, University of Electronic Science and Technology of China}}
\affiliation[7]{organization={School of Information and Software Engineering, University of Electronic Science and Technology of China}}

\begin{abstract}
Continual learning has emerged as a pivotal area of research, primarily due to its advantageous characteristic that allows models to persistently acquire and retain information. However, catastrophic forgetting can severely impair model performance. In this study, we address network forgetting by introducing a novel framework termed Optimally-Weighted Maximum Mean Discrepancy (OWMMD), which imposes penalties on representation alterations via a Multi-Level Feature Matching Mechanism (MLFMM). Furthermore, we propose an Adaptive Regularization Optimization (ARO) strategy to refine the adaptive weight vectors, which autonomously assess the significance of each feature layer throughout the optimization process, The proposed ARO approach can relieve the over-regularization problem and promote the future task learning. We conduct a comprehensive series of experiments, benchmarking our proposed method against several established baselines. The empirical findings indicate that our approach achieves state-of-the-art performance.
\end{abstract}


\begin{highlights}
\item An innovative framework termed Optimally Weighted Maximum Mean Discrepancy (OWMMD) is proposed to mitigate catastrophic forgetting in continual learning paradigms.
\item A Multi-Level Feature Matching Mechanism (MLFMM) is pro impose penalties on the modification of feature representations across various tasks
\item An Adaptive Regularization Optimization (ARO) framework that enables the model to evaluate the significance of each feature layer in real-time throughout the optimization process.
\end{highlights}

\begin{keyword}
Continual Learning, Multi-Level Feature Matching, Adaptive Regularization Optimization
\end{keyword}

\end{frontmatter}


\setlength{\intextsep}{5pt plus 1pt minus 1pt}
\setlength{\textfloatsep}{5pt plus 1pt minus 1pt}

\section{Introduction}
\label{intro}
Continual learning (CL) seeks to empower a model to assimilate knowledge from an ongoing influx of data, where new tasks are presented in a sequential manner \cite{Lifelong_sentiment,AComprehensiveSurvey,LearningToPrompt,cl_review,Lifelonginterpretable}. A primary obstacle in CL is the phenomenon of catastrophic forgetting \cite{LifeLong_review, french1999catastrophic, YE3,PARISI201954,wickramasinghe2023continual}, which manifests when a model loses previously acquired knowledge while adapting to new tasks. This issue is particularly exacerbated in contexts where the model encounters limited data \cite{shin2017continual} for each task and lacks the opportunity to revisit earlier data. Various approaches have been proposed to alleviate catastrophic forgetting \cite{wang2022learning,cossu2021continual}, encompassing rehearsal-based techniques, regularization strategies, and architectural innovations.

Recent research has introduced various methodologies to tackle the issue of network forgetting in continual learning, which can be categorized into three primary approaches: memory-based methods \cite{DER,A-GEM,GEM}, dynamic expansion-based methods \cite{NoSelection,Lifelong_expandable}, and regularization-based methods \cite{CL_Bayesian,BayesianCL}. Among these, the rehearsal-based strategy is a straightforward yet effective technique to mitigate network forgetting, focusing on preserving a subset of historical data. These stored samples are reutilized and integrated with new data to optimize the model during the learning of new tasks \cite{RainbowMemory, YE2}. However, the efficacy of rehearsal-based methods is heavily dependent on the quality of the retained samples. Regularization-based techniques \cite{EWC,si} aim to prevent significant changes to crucial model parameters by incorporating additional regularization terms into the main objective function. Despite this, regularization-based methods primarily focus on weight conservation and may struggle with the diverse and evolving nature of data encountered in real-world scenarios. Recent advancements have proposed employing dynamic expansion models to address network forgetting in continual learning \cite{Lifelonginterpretable}. These methods dynamically generate new sub-networks and hidden layers to assimilate new information over time while preserving all previously trained parameters to ensure optimal performance on earlier tasks. Nonetheless, a notable limitation of the dynamic expansion model is the increasing complexity of the model as the number of tasks expands.

Recent advancements in continual learning have increasingly emphasized knowledge distillation-based methodologies \cite{gou2021knowledge,hinton2015distilling,romero2014fitnets}, wherein the fundamental principle involves transferring knowledge from a previously trained model to the model currently being trained on the active task. In this context, the knowledge distillation framework typically comprises a teacher module, which is trained on the prior task and subsequently frozen, alongside a student module that undergoes continuous training on the current task. A regularization term is incorporated to minimize the divergence between the outputs of the teacher and student, thereby regulating the model's optimization process to mitigate the risk of catastrophic forgetting. The efficacy of knowledge distillation methods is demonstrated by their capacity to maintain performance on earlier tasks without necessitating the retention of any memorized samples, establishing them as a prominent strategy for addressing challenges in continual learning. Nonetheless, a significant limitation of knowledge distillation approaches is their propensity to impair the learning capacity for new tasks, leading to the over-regularization problem.

In this paper, we tackle network forgetting in continual learning by proposing an innovative framework known as Optimally-Weighted Maximum Mean Discrepancy (OWMMD). This framework is designed to mitigate representation shifts through a probability-based distance measure. Specifically, a novel Multi-Level Feature Matching Mechanism (MLFMM) is proposed to minimize the distance between previously and currently learned representations throughout the model's optimization process, effectively addressing the challenge of network forgetting. Unlike conventional distillation techniques that concentrate on aligning final outputs, MLFMM penalizes alterations in representations across all multi-level feature layers, which can further relieve network forgetting. To address the problem of over-regularization, OWMMD incorporates a new Adaptive Regularization Optimization (ARO) strategy that automatically assesses the significance of each feature layer during the model optimization process, thereby preventing network forgetting while facilitating the learning of future tasks. We conduct a comprehensive series of experiments and benchmark our method against several contemporary baselines. The empirical findings indicate that our proposed approach achieves state-of-the-art performance.

We summarize our main contributions in the following:

\begin{enumerate}
    \item We propose a framework termed Optimally-Weighted Maximum Mean Discrepancy (OWMMD) aimed at mitigating catastrophic forgetting in continual learning paradigms. The OWMMD framework utilizes a Multi-Level Feature Matching Mechanism (MLFMM) to impose penalties on the modification of feature representations across various tasks.
    \item To address over-regularization issues caused by the MLFMM, we propose to evaluate the significance of each feature layer when calculating the regularization loss term we call our approach as Adaptive Regularization Optimization (ARO). This approach can adaptively penalize significant changes on important feature layers while encouraging other feature layers to adapt to new tasks. 
    \item We perform comprehensive experiments to assess the efficacy of our methodology, benchmarking it against multiple recognized baselines in continual learning. The findings indicate that our approach achieves state-of-the-art performance, successfully alleviating the problem of forgetting while preserving high accuracy.
\end{enumerate}

The subsequent sections of this manuscript are organized as follows: Chapter 2 provides an overview of the Related Work in continual learning, critically evaluating the advantages and drawbacks of current methodologies. Chapter 3 outlines the Methodology, elaborating on the OWMMD framework and the adaptive regularization optimization component. In Chapter 4, we detail the experimental design and outcomes, showcasing the efficacy of our approach across multiple continual learning benchmarks. Lastly, Chapter 5 wraps up the paper with a synthesis of the findings and suggestions for future research avenues.

\section{Related Work}
\label{relatedWork}

Various methods have been proposed to tackle the challenge of catastrophic forgetting, often categorizing into approaches such as rehearsal-based methods, knowledge distillation, regularization-based methods, and architecture-based methods. We summarize many important baselines in \tablename~\ref{tab:cl_methods}.

\noindent
\textbf{Rehearsal-based} methods store a small subset of past data, often referred to as an episodic memory, and replay this data during training on new tasks to retain knowledge from previous tasks \cite{DER,A-GEM,GEM}.
ER \cite{ER} demonstrates that even very small memory buffers can significantly improve generalization. The study found that training with just one example per class from previous tasks led to improvements about 10\% performance across various benchmarks, suggesting that simple memory-based methods can outperform more complex continual learning algorithms.
GEM \cite{GEM} mitigates forgetting by constraining the optimization process to avoid interference with previous tasks. GEM allows for beneficial transfer of knowledge to new tasks, showing strong performance on datasets such as MNIST \cite{mnist,MNIST2} and CIFAR-100 \cite{sharma2018analysis}.
A-GEM \cite{A-GEM} builds on GEM but improves its computational and memory efficiency, providing a better trade-off between performance and resource consumption. A-GEM also demonstrated the ability to learn more efficiently when provided with task descriptors.
GSS \cite{GSS} proposes an efficient way to select samples for replay buffers by maximizing the diversity of samples. This method frames sample selection as a constraint reduction problem, where the goal is to choose a fixed subset of constraints that best approximate the feasible region defined by the original constraints. This approach outperforms traditional task-boundary-based methods in terms of accuracy and efficiency.
HAL \cite{HAL} introduces a bilevel optimization technique that complements experience replay by anchoring the knowledge from past tasks. By preserving predictions on anchor points through fine-tuning on episodic memory, this method improves both accuracy and the mitigation of forgetting compared to standard experience replay.
For industrial applications, TRINA \cite{TRINA} introduces a federated continual learning framework with self-challenge rehearsal, which generates historical distributions through masked recovery tasks using random scales and positions. This method enhances the model’s ability to recall complex data distributions while maintaining training stability in industrial monitoring scenarios with spatiotemporal heterogeneity.

\noindent
\textbf{Knowledge Distillation} has also been widely adopted in CL \cite{CLTeacherStudent, KDGAN} to combat forgetting by transferring knowledge from a "teacher" model (representing prior knowledge) to a "student" model (the current model). 
In Learning Without Forgetting (LWF) \cite{LWF}, a smoothed version of the teacher’s output is used to guide the student’s responses, ensuring that knowledge learned on previous tasks does not degrade.
iCaRL \cite{icarl} addresses class-incremental learning, where the model is required to learn from a continuously growing set of classes without access to previous data. iCaRL simultaneously learns both strong classifiers and data representations, making it suitable for deep learning architectures. The method uses knowledge distillation to retain knowledge from previous classes while adding new ones. Experiments on CIFAR-100 and ImageNet datasets show that iCaRL can effectively learn a large number of classes incrementally without forgetting previously learned ones, outperforming other methods that rely on fixed representations.
Dark Experience Replay (DER) \cite{DER} addresses the challenges in General Continual Learning (GCL), where task boundaries are not clearly defined and the domain and class distributions may shift gradually or suddenly. The method combines rehearsal with knowledge distillation and regularization. DER works by matching the network’s logits sampled throughout the optimization trajectory, thereby promoting consistency with its past outputs.
In the work \cite{szatkowski2024adapt}, authors investigate exemplar-free class incremental learning (CIL) using knowledge distillation (KD) as a regularization strategy to prevent forgetting. The authors introduce Teacher Adaptation (TA), a method that concurrently updates both the teacher and main models during incremental training. This method seamlessly integrates with existing KD-based CIL approaches and provides consistent improvements in performance across multiple exemplar-free CIL benchmarks.

\noindent
\textbf{Regularization-based} methods seek to prevent catastrophic forgetting by restricting how much the model's parameters can change during training \cite{CL_Bayesian,BayesianCL}. EWC \cite{EWC} presents a seminal approach for alleviating forgetting in neural networks. The method involves selective slowing down of learning on weights that are important for previous tasks. By preserving the important weights through a penalty on large weight changes, the model can continue learning new tasks while maintaining knowledge from previous ones. This regularization approach was demonstrated on both classification tasks (MNIST) and sequential learning tasks (Atari 2600 games), showing that the model can successfully retain knowledge even after long periods without encountering the original tasks.
oEWC \cite{oEWC} introduces a regularization strategy within a progress and compress framework. The method works by partitioning the network into a knowledge base (which stores information about previous tasks) and an active column (which focuses on the current task). After a new task is learned, the active column is consolidated into the knowledge base, protecting the knowledge acquired so far. The key aspect of this approach is its ability to achieve this consolidation without growing the architecture, storing old data, or requiring task-specific parameters. It is shown to work effectively on sequential classification tasks and reinforcement learning domains like Atari games and maze navigation, providing a balance between learning new tasks and preserving old knowledge.
SI \cite{si} mimics the adaptability of biological neural networks. In this approach, each synapse accumulates task-relevant information over time, allowing the model to store new memories while maintaining knowledge from prior tasks. The accumulated task-specific information helps the model regulate how much each synapse can change, ensuring that learning of new tasks does not interfere with older knowledge. SI significantly reduces forgetting and provides a computationally efficient method for continual learning, as demonstrated in classification tasks on benchmark datasets like MNIST.
RW \cite{rw} incorporates a KL-divergence-based perspective to measure the difference between the current model and the previous task's knowledge. The method introduces two new metrics, forgetting and intransigence, to quantify how well an algorithm balances retaining old knowledge and updating for new tasks. The results show that RW offers superior performance compared to traditional methods, as it provides a better trade-off between forgetting and intransigence. The approach is evaluated on several benchmark datasets (MNIST, CIFAR-100), with RW showing significant improvement in preserving knowledge while being able to update for new tasks.

 Feature matching techniques have been introduced to mitigate catastrophic forgetting in continual learning, with a representative method outlined in \cite{metaLearningForgetting}. This approach seeks to minimize the optimal transport distance between previous and current feature representations. However, computing optimal transport is often computationally prohibitive, and \cite{metaLearningForgetting} addresses this challenge using the method from \cite{OPDistance}, which incurs significant computational overhead. In contrast, our work leverages Maximum Mean Discrepancy (MMD) for feature alignment, offering a more computationally efficient solution. Moreover, while \cite{metaLearningForgetting} focuses solely on penalizing changes in the final feature outputs and overlooks intermediate representations, our proposed MLFMM framework enforces regularization across all feature layers, effectively alleviating network forgetting in continual learning scenarios. Additionally, we introduce an Adaptive Regularization Optimization (ARO) strategy that selectively constrains key feature layers, thereby preventing over-regularization. In contrast, \cite{metaLearningForgetting} does not have such an adaptive feature matching mechanism and would easily lead to over-regularization issues.

\begin{table}[t]
\centering
\fontsize{9}{11}\selectfont
\renewcommand{\arraystretch}{1.5}
\caption{Summary of Related Works in Continual Learning.}
\label{tab:cl_methods}
\vspace{5pt}
\resizebox{\textwidth}{!}{ 
    \begin{tabular}{@{}l p{6cm} p{5cm}@{}}
        \toprule
        \textbf{Method Type} & \textbf{Description} & \textbf{Representative} \\ 
        \midrule
        \multirow{1}{*}{\textbf{Rehearsal-based}} & Store and replay past data to retain knowledge from previous tasks. & ER \cite{ER}, GEM \cite{GEM}, A-GEM \cite{A-GEM}, GSS \cite{GSS}, HAL \cite{HAL}, TRINA \cite{TRINA} \\
        \multirow{1}{*}{\textbf{Knowledge Distillation}} & Transfer knowledge from a teacher model to a student model to prevent forgetting. & LWF \cite{LWF}, iCaRL \cite{icarl}, DER \cite{rmkd} \\
        \multirow{1}{*}{\textbf{Regularization-based}} & Restrict changes to model parameters to avoid catastrophic forgetting. & EWC \cite{EWC}, oEWC \cite{oEWC}, SI \cite{si}, RW \cite{rw} \\
        \multirow{1}{*}{\textbf{Architecture-based}} & Alter network structure to accommodate new tasks. & HAT \cite{HAT}, DualNets \cite{fastslow}, MoE \cite{moe}, DBLF \cite{DBLF}, knowledge-guided prompt \cite{kgp}, multi-LoRA \cite{mlora}, WKNN-CLCMTVD \cite{WKNN-CLCMTVD}, \cite{SVOBODA}\\ 
        \bottomrule
    \end{tabular}}
\end{table}

\noindent
\textbf{Architecture-based} approaches tackle catastrophic forgetting by altering the network structure to accommodate new tasks \cite{NoSelection,Lifelong_expandable,YE2,dytox,YE4,Expertgate}. 
HAT \cite{HAT} proposes a task-based hard attention mechanism to prevent catastrophic forgetting. In this approach, a hard attention mask is learned for each task during training, allowing the network to focus on the relevant parameters for each specific task while ignoring irrelevant ones. This prevents forgetting by restricting updates to previously learned tasks. The hard attention mask is updated during training, with previous masks conditioning the learning of new tasks. This method significantly reduces forgetting and provides flexibility in controlling the stability and compactness of learned knowledge, making it suitable for both online learning and network compression applications.
Quang Pham et al \cite{fastslow} introduce a DualNets framework based on the Complementary Learning Systems (CLS) theory from neuroscience, which posits that human learning occurs through two complementary systems: a fast learning system for individual experiences (hippocampus) and a slow learning system for structured knowledge (neocortex). DualNets implements this idea in deep neural networks by dividing the network into two components: a fast learning system for supervised learning of task-specific representations and a slow learning system for learning task-agnostic, general representations through Self-Supervised Learning (SSL). This dual approach helps the model balance learning efficiency with retention of previously learned knowledge, making it effective in both task-aware and task-free continual learning scenarios. DualNets has shown strong performance on benchmarks like CTrL, outperforming dynamic architecture methods in some cases.
Hongbo et al explore the MoE \cite{moe} architecture in the context of continual learning, providing the first theoretical analysis of MoE's impact in this domain. MoE models use a collection of specialized experts, with a router selecting the most appropriate expert for each task. The paper shows that MoE can diversify its experts to handle different tasks and balance the workload across experts. The study suggests that MoE in continual learning may require termination of updates to the gating network after sufficient training rounds for system convergence, a condition that is not required in non-continual MoE studies. Additionally, the paper provides insights into expected forgetting and generalization error in MoE, highlighting that adding more experts can delay convergence without improving performance. The theoretical insights are validated through experiments on both synthetic and real-world datasets, demonstrating the potential benefits of MoE in continual learning for deep neural networks (DNNs).
DBLF \cite{DBLF} constructs dedicated branch layers for old tasks and dynamically fuses them with new task layers via a two-stage training process (adaptation and fusion), effectively addressing model growth in imbalanced rotating machinery fault diagnosis. 
Lu et al \cite{kgp} introduce knowledge-guided prompt alignment with contrastive hard negatives, improving task-specific prompt distinguishability through a semantic-enhanced module. 
For universal information extraction, Jin et al designed a multi-LoRA architecture \cite{mlora} that freezes pretrained weights and trains independent low-rank adapters (LoRA) for auxiliary and target tasks, merging parameters during inference. 
WKNN-CLCMTVD \cite{WKNN-CLCMTVD} continuously updates bio-inspired memory cells using k-nearest neighbor rules, enabling fast adaptation to time-varying data spaces.
Svoboda et al \cite{SVOBODA} combines Hoeffding trees with PELT-based change point detection to dynamically select model ensembles for non-stationary natural gas consumption forecasting.

\section{Methodology}
\label{method}

\subsection{Problem definition}
The description of mathematical notations of this paper is showed in \tablename~\ref{tab:notion}. In continual learning, a model is unable to access the complete training dataset simultaneously and instead acquires knowledge from a continuously evolving data stream. A common scenario in continual learning is referred to as class-incremental learning, which encompasses a sequence of $N$ tasks $ \{T_1, T_2, \dots, T_N\}$. Each task $T_i$ is linked to a labeled training dataset $ D^s_i = \{({\bf x}^i_{j}, {\bf y}^i_{j}) \,|\, j = 1,\cdots,N^s_{i} \} $, where $ {\bf x}^i_j \subset \mathcal{X} $ and $ {\bf y}^i_j \subset \mathcal{Y}$ represent the $j$-th paired sample from the $i$-th task, with $N^s_{i}$ indicating the total number of samples in the training dataset $D^s_i$. Here, $\mathcal{X}$ and $\mathcal{Y}$ denote the respective spaces of data samples and labels. Additionally, let $ D^t_i = \{({\bf x}^i_j, {\bf y}^i_j) \,|\, j = 1,\cdots,N^t_i \} $ represent the testing dataset for the $i$-th task, where $N^t_i$ signifies the total number of testing samples associated with the $i$-th task.

In a continual learning framework, the goal of the model is to identify the optimal parameter set ${\theta^\star}$ from the parameter space $\tilde{\Theta}$, which minimizes the training loss across all tasks $\{{T_1, \cdots, T_i}\}$ while learning the current task $T_i$, as expressed by the following equation~:
\begin{equation}
    \begin{aligned}
    \theta^\star = \underset{\theta \in \tilde{\Theta} }{\operatorname{argmin}} 
    \frac{1}{i} \sum^{i}_{k=1} 
    \Big\{
     \frac{1}{N^s_i}
    \sum^{N^s_i}_{j=1}
    \big\{
    {\mathcal{L}}
    \left({\bf y}^k_j, f_{\theta}({\bf x}^k_j) \right) \big\}  \Big\} ,
    \label{eq:problem}
    \end{aligned}
\end{equation}
where $\theta^\star$ denotes the optimal model parameters, while ${\mathcal{L}}(\cdot,\cdot)$ represents a loss function that can be realized through the cross-entropy loss. Additionally, $f_\theta(\cdot) \colon \mathcal{X} \to \mathcal{Y}$ functions as a classifier that receives ${\bf x}^i_j$ as input and produces the corresponding predicted label.

Identifying the optimal parameter $\theta^\star$ through Eq.~\eqref{eq:problem} in the context of continual learning presents significant challenges, as the model is restricted to utilizing only the data samples from the current task ($T_i$), while all preceding tasks $\{ T_1,\cdots,T_{i-1} \}$ remain inaccessible. Research in continual learning seeks to devise various methodologies aimed at determining the optimal parameter that effectively minimizes the training loss across all tasks. Upon the completion of the final task ($T_N$), we assess the model's performance against all testing datasets $\{ D^t_1,\cdots,D^t_N \}$.

\begin{table}[ht]
\centering
\fontsize{9}{11}\selectfont
\renewcommand{\arraystretch}{1.1} 
\caption{Description of Mathematical Notations}
\vspace{5pt}
\label{tab:notion}
\begin{tabular}{@{}l l@{}} 
\toprule 
\textbf{Notation} & \textbf{Description} \\
\midrule
$T_i$, $B_i$ & The $i$-th task in a sequence $\{T_1, T_2, \dots, T_N\}$ and number of batches of $T_i$. \\
$D^s_i, D^t_i$ & Training/Testing datasets for task $T_i$: $D^s_i = \{ {\bf x}^i, {\bf y}^i \}$, $D^t_i = \{ {\bf x}^{t,i}, {\bf y}^{t,i} \}$. \\
$N^s_i$, $N^t_i$ & Number of samples in $D^s_i$ and $D^t_i$. \\
$\mathcal{X}, \mathcal{Y}$ & Space of data samples and labels: $\mathbf{x}^i \subset \mathcal{X}$, $\mathbf{y}^i \subset \mathcal{Y}$. \\
$\tilde{\Theta}, \theta^\star$ & Set of model parameters and the optimal set found via optimization. \\
$\theta_i$ & The parameter of the model in $T_i$. \\
$F_{\theta_i}$ & The model with the parameter $\theta_i$. \\
$F_{\theta_i, k}$ & The $k$-th feature layer of $F_{\theta_i}$. \\
$G_{\theta_i}$ & Linear classifier of $F_{\theta_i}$. \\
${\mathcal{Z}}^k$ & Feature space of the outputs of $F_{\theta_i, k}$. \\
$H_{\theta_i}$ & Logits output: $H_{\theta_i}(\mathbf{x}) = G_{\theta_i}(F_{\theta_i,K}(\cdots F_{\theta_i,1}(\mathbf{x})))$ \\
$\mathcal{Z}^c$ & Output space of classifier $G_{\theta_i}$ (logits space) \\
$\mathcal{F}({\theta_i},\cdot,k)$ & The feature vector extracted by $F_{\theta_i, k}$. \\
$f_\theta(\cdot)$ & Classifier mapping $\mathcal{X}$ to $\mathcal{Y}$. \\
$\mathcal{L}_s(\cdot,\cdot)$ & Cross-entropy loss function. \\
$\mathcal{L}_r$ & Multi-level regularization term using MMD: $\sum_{k=1}^K \tilde{w}_k \mathcal{L}^e_{\rm M}(P_{\mathbf{Z}^k_i}, P_{\mathbf{Z}^k_{i+1}})$ \\
$\mathcal{L}^e_{\rm M}(\cdot,\cdot)$ & Unbiased MMD estimator between feature distributions (Eq. \ref{MMD_criterion5}) \\
$\mathcal{M}_i$ & Memory buffer of $T_i$. \\
$\alpha, \beta, \gamma$ & Coefficients that balance the importance of each term in the loss function \\
$w_k, \tilde{w}_k$ & Adaptive weight for layer $k$ and its softmax-normalized version: $\tilde{w}_k = e^{w_k}/\sum_j e^{w_j}$ \\
\bottomrule 
\end{tabular}
\end{table}

\subsection{Multi-Level Feature Matching Mechanism}

In the realm of continual learning, numerous studies have advocated for the utilization of knowledge distillation \cite{han2022online,varkd,szatkowski2024adapt} methodologies to mitigate the phenomenon of network forgetting. The core principle of knowledge distillation involves maintaining the previous model as a teacher module and synchronizing the outputs between the teacher and the student module that is represented by the current active classifier. This alignment seeks to minimize substantial alterations in critical parameters of the current model (student) during the acquisition of new tasks \cite{co2l}. Nevertheless, the majority of existing knowledge distillation techniques primarily focus on deriving a regularization term within the prediction space, often neglecting the semantic feature space. In this paper, we present an innovative approach termed the Multi-Level Feature Matching Mechanism (MLFMM), designed to regulate information flow within the feature space, which can effectively avert network forgetting.

Let us formally define the function $F_{\theta_i} \colon {\mathcal{X}}  \to {\mathcal{Y}}$ as a model parameterized by $\theta_i$, which comprises $K$ feature layers denoted as $\{ F_{\theta_i,1},F_{\theta_i,2},\cdots,F_{\theta_i,K}  \}$, along with a linear classifier represented by $G_{\theta_i} \colon {\mathcal{Z}}  \to {\mathcal{Y}}$. Here, the index $i$ signifies the model's adaptation during the $i$-th task learning phase, and ${\mathcal{Z}}$ refers to the feature space corresponding to the output of the feature layer. For a specified input ${\bf x}$, we can derive the feature vector from a particular feature layer through the following process~:

\begin{equation}
    \begin{aligned}
        \mathcal{F}({\theta_i},{\bf x},k) =  F_{\theta_i,k} \left( F_{\theta_i,k-1} \left( \cdots F_{\theta_i,1} ({\bf x}) \right) \right)\,,
    \end{aligned}
\end{equation}
\noindent where each feature layer $F_{\theta_i,k}$ receives the output from the last layer $F_{\theta_i,k-1}$ and produces the feature vector over the space ${\mathcal{Z}}^k$. The output of the final feature layer, denoted as ${\bf z} \colon {\bf z} \in {\mathcal Z}$, is computed by applying the linear transformations~:
\begin{equation}
\begin{aligned}
    {\bf z} = F_{\theta_i,K}(F_{\theta_i,K-1}( \cdots F_{\theta_i,1}({\bf x}))) \,,
    \label{eq:logits}
    \end{aligned}
\end{equation}
The whole prediction for the model $F_{\theta_i}$ can be expressed as~
\begin{equation}
\begin{aligned}
    {\bf \hat{y}} = Softmax(G_{\theta_i}({\bf z})) \,,
\end{aligned}
\end{equation}
\noindent where ${\bf \hat{y}}$ is the prediction for the input ${\bf x}$. The $Softmax()$ is the normalized function.
In this paper, we adopt the cross-entropy loss function to update the parameter ${\theta_i}$, defined by~:
\begin{equation}
\begin{aligned} 
{\mathcal{L}}_{s} ( {\bf y}, {\bf \hat{y}} ) =   \sum_{t=1}^C \big\{ {\bf y}[t] \log ( {\bf \hat{y}}[t] ) \big\}  \,,
\label{crossEntropy}
\end{aligned}
\end{equation}
\noindent where $C$ represents the total number of categories. ${\bf y}[t]$ and ${\bf \hat{y}}[t]$ denote the $t$-th dimension of the real class label and the corresponding prediction.

To mitigate the issue of network forgetting, this study implements a memory buffer ${\mathcal{M}}_i$ to retain a limited number of historical examples. In particular, we propose utilizing Reservoir Sampling for the sample selection process, which offers computational efficiency. Nonetheless, the model's performance on prior tasks diminishes when the memory capacity is constrained. To tackle this challenge, the proposed OWMMD regulates the updates to the model's representations during the learning of new tasks. Specifically, upon the completion of the current task $T_i$, we preserve and freeze the parameters of all feature layers $\{F_{\theta_i,1},\cdots,F_{\theta_i,K} \}$ during the subsequent task learning $T_{i+1}$. A regularization term is incorporated to minimize the discrepancy in representations between the previous model ${F_{\theta_i}}$ and the current model ${F_{\theta_{i+1}}}$ at $T_{i+1}$~:
\begin{equation}
\begin{aligned}
{\mathcal{F}}_{r} ({\theta_i},{\theta_{i+1}}, {\bf x}) = \sum_{k=1}^{K} \big\{ F_d({\mathcal F}({\theta_i},{\bf x},k), {\mathcal F}({\theta_{i+1}},{\bf x},k) \big\}  \,,
\label{eq:regulation}
\end{aligned}
\end{equation}
\noindent where $F_d(\cdot,\cdot)$ is a distance measure function and $K$ is the total number of feature layers. In the following section, we introduce a probabilistic distance to implement Eq.~\eqref{eq:regulation}.

\subsection{The Maximum Mean Discrepancy based Regularization}

Maximum Mean Discrepancy (MMD) \cite{MMD_criteria} serves as a significant and widely utilized distance metric within the realm of machine learning, grounded in a kernel-based statistical framework designed to assess the equivalence of two data distributions. Owing to its robust distance estimation capabilities, MMD has been adopted as a fundamental loss function for training diverse models across various applications, including image synthesis and density estimation. In contrast to alternative distance metrics such as Kullback–Leibler (KL) divergence and Earth Mover's Distance, the principal advantage of the MMD criterion lies in its incorporation of the kernel trick, which facilitates the estimation of MMD on vectors without necessitating knowledge of the specific form of the density function.

The Maximum Mean Discrepancy (MMD) criterion is predicated on the concept of embedding probability distributions within a Reproducing Kernel Hilbert Space (RKHS). Let us denote $A$ and $B$ as two Borel probability measures. We introduce ${\bf a}$ and ${\bf b}$ as random variables defined over a topological space $\mathcal{X}$. Furthermore, we characterize the set $\{ f \in \mathcal{F} | f \colon \mathcal{X} \to \mathbf{R}\}$ as a function, where $\mathcal{F}$ represents a specific class of functions. The MMD criterion quantifying the divergence between the distributions ${A}$ and ${B}$ is articulated as \cite{MMD_criteria}~:
\begin{equation}
\begin{aligned}
{\mathcal{L}}_{\rm M}(A, B) \buildrel \Delta \over = \mathop {\sup }\limits_{f \in {\mathcal{F}}} \left( {{{\mathbb{E}}_{{\bf a} \sim A }}\left[ {f({\bf a})} \right] - {{\mathbb{E}}_{{\bf b }\sim B }}\left[ {f( {\bf b})} \right]} \right)\,.
\label{MMD_criterion}
\end{aligned}
\end{equation}
where $\rm sup(\cdot)$ indicates the least upper bound of a set of numbers. 
If two distribution are same $A=B$, we can have ${\mathcal{L}}_{\rm M}(A, B) = 0$. The function class $\mathcal{F}$ is implemented as a unit ball in an RKHS with a positive definite kernel $f_k({\bf a},{\bf a}')$. Eq.~\eqref{MMD_criterion} is usually hard to be calculated, it can be estimated on the embedding space \cite{MMDGAN}, which is defined as~:
\begin{equation}
    \begin{aligned}
        {\mathcal{L}}^2_{{\rm M}}(A, B) &=   || {\boldsymbol{\mu}}_{A} - {\boldsymbol{\mu}}_{B} ||^2\,,
        \label{mmd2_eq}
    \end{aligned}
\end{equation}
where ${\boldsymbol{\mu}}_{A}$ and ${\boldsymbol{\mu}}_{B}$ are the mean embedding of the two distributions $A$ and $B$, respectively. $||\cdot||^2$ is the Euclidean distance. Each mean embedding ${\boldsymbol{\mu}}_{A}$ is defined as~:
\begin{equation}
    \begin{aligned}
    {\boldsymbol{\mu}}_{A} = \int f_k({\bf a},\cdot) \frac{\partial P({\bf a})}{\partial {\bf a}} d{\bf a}\,,
    \end{aligned}
\end{equation}
where $P({\bf a})$ is the probability density function for the distributioin $A$. Each mean embedding ${\boldsymbol{\mu}}_{P}$ also satisfies the following equation~:
\begin{equation}
    \begin{aligned}
    {\mathbb{E}}[ f({\bf a}) ] = \left\langle f,  {\boldsymbol{\mu}}_{A} \right\rangle_{\mathcal{H}} \,,
    \end{aligned}
\end{equation}
where $\left\langle {f} \right.,\left. \cdot \right\rangle_{\mathcal{H}}$ is the inner product. Specifically due to the reproducing property of RKHS $f \in {\mathcal{F}}, f( {\bf a}) =  \left\langle {f} \right.,\left. {f_k( {\bf a},\cdot)} \right\rangle_{\mathcal{H}}$, we can solve Eq.~\eqref{mmd2_eq} by considering the kernel functions~:
\begin{equation}
\begin{aligned}
\mathcal{L}^2_{\rm M}(A, B) = \mathbb{E}_{{\bf a},{\bf a}' \sim A}[f_k({\bf a},{\bf a}')]
- 2\mathbb{E}_{{\bf a} \sim P, {\bf b} \sim B}[f_k({\bf a},{\bf b})] + \mathbb{E}_{{\bf b},{\bf b}' \sim B}[f_k({\bf b},{\bf b}')]\,,
\label{MMD_criterion3}
\end{aligned}
\end{equation}
where ${\bf a}'$ and ${\bf b}'$ denotes independent copies of the samples ${\bf a}$ and ${\bf b}$, respectively. In practice, we usually collect the same number of samples from the two distributions $A$ and $B$ ($N_A = N_B = N$), where $N_A$ and $N_B$ denote the number of samples from two distributions $A$ and $B$, respectively. We can estimate Eq.~\eqref{MMD_criterion3} by considering an unbiased empirical measure~:
\begin{equation}
\begin{aligned}
{\mathcal L}^e_{\rm M}(A, B) &= \frac{1}{N(N-1)} \sum_{i=1}^{N}\sum_{i \ne j} \Big\{ h(i,j) \Big\} \,,
\label{MMD_criterion5}
\end{aligned}
\end{equation}
where $h(i,j) = f_k({\bf a}_i,{\bf a}_j) + f_k({\bf b}_i,{\bf b}_j)- f_k({\bf a}_i,{\bf b}_j) - f_k({\bf a}_j,{\bf b}_i)$. To apply Eq.~\eqref{MMD_criterion5} for the proposed regularization term defined in Eq.~\eqref{eq:regulation}, we first from a set of feature vectors, expressed as~:
\begin{equation}
\begin{aligned}
    {\bf Z}_i = \{ {\bf Z}_i^1,\cdots,{\bf Z}_i^K \}\,,
    \end{aligned}
\end{equation}
\noindent where each ${\bf Z}^{j}_i$ is a feature vector formed using a batch of data samples $\{ {\bf x}_1,\cdots,{\bf x}_b\}$, expressed as~:
\begin{equation}
\begin{aligned}
{\bf Z}^{j}_i = \{ \mathcal{F}({\theta_i},{\bf x}_1,j),\cdots,\mathcal{F}({\theta_i},{\bf x}_b,j)  \}\,,
\end{aligned}
\end{equation}
Similar, we can form the feature vector ${\bf Z}^j_{i+1}$ using the current model $F_{\theta_{i+1}}$ on the data batch $\{ {\bf x}_1,\cdots,{\bf x}_b \}$, expressed as~:
\begin{equation}
\begin{aligned}
{\bf Z}^{j}_{i+1} = \{ \mathcal{F}({\theta_{i+1}},{\bf x}_1,j),\cdots,\mathcal{F}({\theta_{i+1}},{\bf x}_b,j)  \}\,.
\end{aligned}
\end{equation}
Let $P_{{\bf Z}^j_i}$ and $P_{{\bf Z}^j_{i+1}}$ denote the distribution of ${\bf Z}^j_i$ and ${\bf Z}^j_{i+1}$, respectively. Based on the MMD criterion, we can implement the regularization function when seeing the data batch $\{ {\bf x}_1,\cdots,{\bf x}_b \}$ at the new task learning $T_{i+1}$ as~:
\begin{equation}
    \begin{aligned}
{\mathcal{L}}_{r} = \sum_{t=1}^{K} \big\{ 
{\mathcal{L}}^e_{{\rm M}}( P_{{\bf Z}^t_i}, P_{{\bf Z}^t_{i+1}} )
\big\}\,.
\label{eq:regularization}
    \end{aligned}
\end{equation}

Furthermore, we propose utilizing the DER++ \cite{DER} as our foundational model while incorporating the suggested regularization term into the primary objective function. The DER++ framework integrates rehearsal, knowledge distillation, and regularization techniques to effectively tackle General Continual Learning (GCL) challenges \cite{gao2023unified,li2022learning}. This approach ensures that the predictions of the current model are aligned with those from previous tasks through a composite of loss components: the conventional loss associated with the current task, a regularization term, and a distillation term~:
\begin{equation}
\begin{aligned}
    {\mathbb E}_{({\bf x},{\bf y} ) \sim P_{D^s_{i+1}}}[ {\mathcal{L}}_{s} ( {\bf y}, F_{\theta_{i+1}}({\bf x}) )] 
    + \alpha  \mathbb{E}_{({\bf x}, {\bf u} )\sim { P}_{{\mathcal{M}}_{i+1}}} [||{\bf u} - F_{\theta_{i+1}}({\bf x})||^2] \\
    + \beta  \mathbb{E}_{({\bf x}, {\bf y} )\sim { P}_{{\mathcal{M}}_{i+1}}}[\mathcal{L}_s({\bf y}, F_{\theta_{i+1}}({\bf x}))]
\end{aligned}
\end{equation}
\noindent where $P_{D^s_{i+1}}$ and $P_{{\mathcal{M}}_{i+1}}$ denote the distribution of $D^s_{i+1}$ and the memory buffer ${\mathcal{M}}_{i+1}$ at the $(i+1)$-th task learning, ${\bf u}$ is the logits of previous task and was saved in the memory buffer. This approach helps the model retain past knowledge while learning new tasks efficiently.

Since the proposed approach can be smoothly applied to the existing continual leanring models, we consider applying our approach to DER++. The final objective function, including a regularization term $\mathcal{L}_r$, is defined as~:
\begin{equation}
\begin{aligned}
    {\mathcal{L}}_{\rm total} &= {\mathbb E}_{({\bf x},{\bf y} ) \sim P_{D^s_{i+1}}}[ {\mathcal{L}}_{s} ( {\bf y}, F_{\theta_{i+1}}({\bf x}) )] 
    + \alpha  \mathbb{E}_{({\bf x}, {\bf z} )\sim { P}_{{\mathcal{M}}_{i+1}}} [||{\bf u} - F_{\theta_{i+1}}({\bf x})||^2] \\ 
    &+ \beta  \mathbb{E}_{({\bf x}, {\bf y} )\sim { P}_{{\mathcal{M}}_{i+1}}}[\mathcal{L}_s({\bf y}, F_{\theta_{i+1}}({\bf x}))]
    + \gamma {\mathcal{L}}_r \,,
\end{aligned}
\end{equation}
\noindent where $\gamma$ is a hyperparameter to control the importance of the regularization term.

\subsection{Adaptive Regularization Optimization Term}
\label{sec:aw}

To improve the performance of the proposed MMD-based regularization method \cite{MMD_criteria, mmd}, it is essential to automatically allocate distinct weights to each regularization term.
This strategy enables the assignment of varying degrees of significance to the features derived from different layers, thereby accurately reflecting their contributions to the overall learning framework.
Specifically, we can establish an adaptive weight for each layer to assess its relevance to the entire learning process. To ensure that the adaptive weights across all layers are normalized and suitably balanced, we incorporate the softmax function.

Let ${\bf w} = \{w_1,\cdots,w_K \}$ be a trainable adaptive weight vector, where $w_j$ determines the importance of the $j$-th feature layer. To avoid numerical overflow, we propose to normalize all weights using the following equation~:
\begin{equation}
    \begin{aligned}
    \tilde{w}_j = \frac{e^{w_j}} {\sum^K_{{k=1}}e^{w_k}}
    \label{eq:softmax}\,,
    \end{aligned}
\end{equation}
\noindent where $\tilde{w}_j$ is the normalized value of the $j$-th adaptive weight $w_j$. By using the normalized adaptive weights, the regularization function can be redefined as~:
\begin{equation}
    \begin{aligned}
{\mathcal{L}}'_{r} = \sum_{t=1}^{K} \big\{  \tilde{w}_j 
{\mathcal{L}}^e_{{\rm M}}( P_{{\bf Z}^t_i}, P_{{\bf Z}^t_{i+1}} )
\big\}\,.
\label{eq:regulationLoss}
    \end{aligned}
\end{equation}
By using the proposed AROT, the final loss function is redefined as~:
\begin{equation}
\begin{aligned}
    \mathcal{L}'_{\rm total} &= {\mathbb E}_{({\bf x},{\bf y} ) \sim P_{D^s_{i+1}}}[ \mathcal{L}_s ( {\bf y}, F_{\theta_{i+1}}({\bf x}) )] 
    + \alpha  \mathbb{E}_{({\bf x}, {\bf z} )\sim { P}_{{\mathcal{M}}_{i+1}}} [||{\bf z} - F_{\theta_{i+1}}({\bf x})||^2] \\ 
    &+ \beta  \mathbb{E}_{({\bf x}, {\bf y} )\sim { P}_{{\mathcal{M}}_{i+1}}}[\mathcal{L}_s({\bf y}, F_{\theta_{i+1}}({\bf x}))]
    + \gamma {\mathcal L}'_r \,,
\end{aligned}
\end{equation}
During the training procedure, each adaptive weight $w_j$ is optimized by~:
\begin{equation}
\begin{aligned}
w_j = w_j - \eta \nabla {\mathcal{L}}'_{\rm total}\,, j = 1,\cdots,K \,. 
\end{aligned}
\end{equation}
\noindent where $\eta$ is the learning rate.

\subsection{Algorithm Implementation}
The model training process of the proposed OWMMD framework follows a sequence of steps designed to incrementally learn from new tasks while mitigating catastrophic forgetting. The process is divided into distinct stages for each task, and the incorporation of adaptive regularization enhances the model's ability to balance knowledge retention and learning of new tasks. The whole training algorithm of the proposed OWMMD framework is summarized into three steps~:

\begin{figure}[t]
    \centering
    \setlength{\abovecaptionskip}{0.05cm}
    \includegraphics[trim=0cm 18.2cm 25.5cm 0cm, clip, width=\textwidth, page=1]{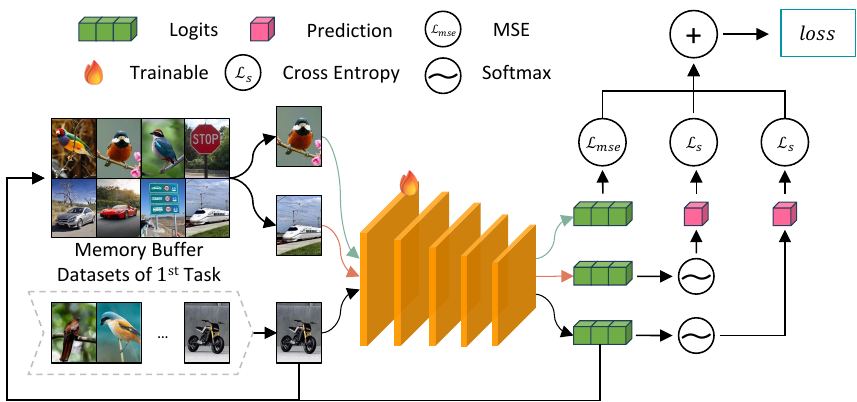}
    \caption{The learning process of the proposed framework during the first task, where the model learns data samples without adaptive regularization. Images of Current Task Dataset and their logits are stored into the buffer via reservoir sampling for future task regularization. Different colored arrows indicate distinct data paths.}
\label{fig:task1}
\end{figure}

\begin{figure}[htp]
    \centering
    \setlength{\abovecaptionskip}{0.05cm}
    \includegraphics[trim=0cm 11cm 25.7cm 0cm, clip, width=\textwidth, page=2]{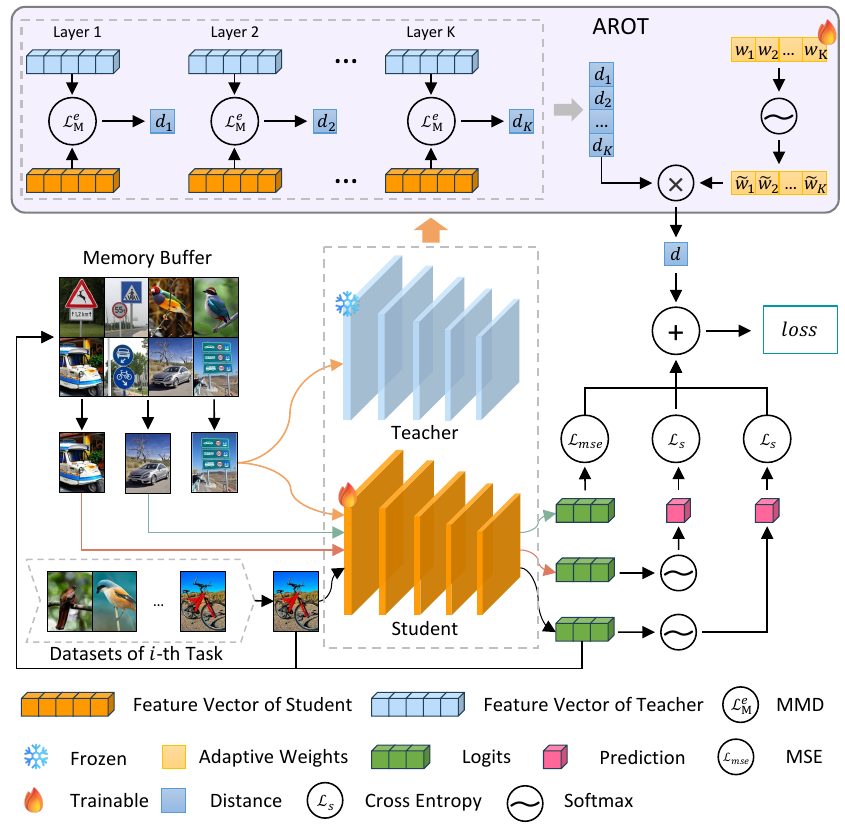}
    \caption{The learning process of the proposed framework during the subsequent tasks, where the adaptive regularization is applied to relieve network forgetting. The regularization term leverages layer-wise Maximum Mean Discrepancy (MMD) between the intermediate features of the teacher network (previous task model) and the student network (current model), guided by adaptive weights to prioritize critical layers. Current task data are processed to generate logits, which are then stored into the buffer via reservoir sampling, ensuring balanced retention of historical knowledge while adapting to new tasks. In the figure, Different colored arrows indicate distinct data paths.}
\label{fig:task2}
\end{figure}

\begin{algorithm}[t]
\caption{The Training Algorithm of the Proposed OWMMD Framework}
\label{alg:owmmd}
\begin{algorithmic}[1]
    \STATE ${\bf Input:}$ dataset $D^s$, parameters $\theta$, scalars $\alpha$, $\beta$ and $\gamma$, learning rate $\eta$, number of model layers $K$
    \STATE ${\bf Init:}$ ${\mathcal M} \leftarrow \{\}$, ${\bf w} \leftarrow Uniform(0,1,K)$
    \FOR{$i = 1$ \TO $N$}
        \STATE $\theta_i \leftarrow \theta$, ${\mathcal M_i} \leftarrow {\mathcal M}$
        \FOR{$({\bf x}, {\bf y})$ in $D^s_i$}
            \STATE $\mathcal{L}_{\rm total} \leftarrow \mathcal{L}_s({\bf y}, F_{\theta_i}({\bf x}))$
            \STATE ${\bf z} \leftarrow H_{\theta_i}({\bf x})$

            \STATE $({\bf x'}, {\bf z'}) \leftarrow sample({\mathcal M_i})$
            \STATE $\mathcal{L}_{\rm total} \leftarrow \mathcal{L}_{\rm total} + \alpha \cdot ||{\bf z'}-H_{\theta_i}({\bf x'})||^2$

            \STATE $({\bf x'}, {\bf y'}) \leftarrow sample({\mathcal M_i})$
            \STATE $\mathcal{L}_{\rm total} \leftarrow \mathcal{L}_{\rm total} + \beta \cdot \mathcal{L}_s({\bf y'}, F_{\theta_i}({\bf x'}))$
            
            \IF{$i$ $>$ 1}                
                \STATE ${\bf x'} \leftarrow sample({\mathcal M_i})$
                \STATE $\mathcal{L}_{\rm total} \leftarrow  \mathcal{L}_{\rm total} + \gamma \cdot {\mathcal L}'_r({\bf x'})$
                \STATE $\theta_i \leftarrow \theta_i - \eta\nabla{\mathcal L}_{\rm total}$
                \STATE ${\bf w} \leftarrow {\bf w} - \eta\nabla{\mathcal L}_{\rm total}$
            \ELSE
                \STATE $\theta_i \leftarrow \theta_i - \eta\nabla{\mathcal L}_{\rm total}$
            \ENDIF
            
            \STATE ${\mathcal M_i} \leftarrow reservoir({\mathcal M}_i, ({\bf x}, {\bf y}, {\bf z}))$
        \ENDFOR
        \STATE $\theta \leftarrow \theta_i$, ${\mathcal M} \leftarrow {\mathcal M_i}$
    \ENDFOR
\end{algorithmic}
\end{algorithm}

\begin{algorithm}[t]
\caption{Adaptive Regularization Optimization Term for $T_i$}
\label{alg:adaptive-regularization}
\begin{algorithmic}[1]
    \STATE ${\bf Input:}$ adaptive weights $\bf w$
    \STATE $w_{sum} \leftarrow 0$
    \FOR{$j = 1$ \TO $c$}
        \STATE $w_{sum} \leftarrow w_{sum}+e^{w_j}$
    \ENDFOR
    \STATE $d \leftarrow 0$
    \STATE ${\bf x'} \leftarrow sample({\mathcal M}) $
    \STATE $\mathcal{F}_t \leftarrow {\bf x'}$
    \STATE $\mathcal{F}_s \leftarrow {\bf x'}$
    \FOR{$k = 1$ \TO $K$}
        \STATE ${\mathcal F}_t, {\mathcal F}_s \leftarrow F_{\theta_{i-1},k}({\mathcal F}_t), F_{\theta_i,k}({\mathcal F}_s)$
        \STATE $\tilde{w}_k \leftarrow e^{w_k}/w_{sum}$
        \STATE $d \leftarrow d + \tilde{w}_k \cdot {\mathcal L}^e_{\rm M}({\mathcal F}_t, {\mathcal F}_s)$
    \ENDFOR
    \STATE ${\bf Output:}$ $d$
\end{algorithmic}
\end{algorithm}

\noindent
\textbf{Step 1: Initialization and Setup.}
The training process begins with initializing the dataset $D^s$, the model parameters $\theta$, and essential hyperparameters $\alpha$, $\beta$, and $\gamma$, which control the respective loss terms. A uniform weight vector $\mathbf{w}$ is also initialized for each layer of the model, and the memory buffer $\mathcal{M}$ is set as an empty set. These initializations provide the foundation for the learning process.

\noindent
\textbf{Step 2: Adaptive Regularization Optimization.}
The model optimizes the adaptive regularization term to ensure the stability of learned features across tasks, as showed in \figurename~\ref{fig:task2}. A key component of this step is the calculation of the distance between the teacher and student models at each layer. Specifically, during training, the teacher model ($\theta_{i-1}$) and the student model ($\theta_i$) are compared using the MMD-based regularization method, which calculates the distance between the features produced by the models at each layer. The detailed layer-wise MMD computation and its integration with adaptive regularization are described in Algorithm~\ref{alg:owmmd}.

The adaptive weights play a critical role in this optimization. Each normalized weight, $\tilde{w}_k$, adjusts the importance of the corresponding feature layer, ensuring that the regularization process focuses on the layers that are most relevant to the current learning task. The softmax-normalized weight vector ensures that the contributions from all layers are appropriately balanced.

Furthermore, the distance between the teacher and student features, denoted as $\mathcal{L}^e_{\rm M}({\mathcal F}_t, {\mathcal F}_s)$, is computed at each layer to relive network forgetting. This mechanism ensures that the student model retains the essential knowledge from previous tasks while adapting to the new data. The final regularization loss integrates these distance terms across all layers, weighted by the adaptive weights.

\noindent
\textbf{Step 3: The Training Process.}
The overall training process is detailed in Algorithm~\ref{alg:adaptive-regularization}. During the first task, as depicted in \figurename~\ref{fig:task1}, the model learns from the initial task without applying the adaptive regularization term. In this phase, the model focuses solely on the current task's loss function. Notably, images from the current task dataset and their corresponding logits are stored into the buffer via reservoir sampling for future regularization. In the figure, different colored arrows indicate distinct data paths.

For subsequent tasks ($i>1$), as shown in \figurename~\ref{fig:task2}, the adaptive regularization term $\mathcal{L'}_r$ is incorporated to mitigate network forgetting. In this stage, the regularization term leverages the layer-wise Maximum Mean Discrepancy (MMD) between the intermediate features of the teacher network (previous task model) and the student network (current model). This process is guided by adaptive weights to prioritize the critical layers. Additionally, current task data are processed to generate logits, which are then stored into the buffer via reservoir sampling, ensuring balanced retention of historical knowledge while adapting to new tasks.

The model continuously trains over all tasks in the sequence, updating both the adaptive weights $\mathbf{w}$ and the model parameters $\theta_i$ during each task learning phase. After each task, the model's performance is evaluated and the memory buffer is updated with new data samples for future use, allowing the model to progressively generalize across tasks while preserving knowledge from previous ones.

\section{Theoretical Framework}
In this section, we build upon the domain adaptation theory \cite{DomainAdaotionTheory} to establish a novel theoretical framework for analyzing our proposed methodology. We begin by introducing key notations and definitions essential for the subsequent analysis.

\vspace{2pt}
\noindent
\textbf{Definition 1.} Let us define $P_{{\bf Z}^j_i}$ and $P_{{\bf Z}^j_{i+1}}$ as the distribution of the feature vectors ${\bf Z}^j_i$ and ${\bf Z}^j_{i+1}$ produced by the $j$-th representation layer of the teacher and student model, respectively. Let $h \in {\mathcal{H}}$ denote a linear classifier and ${\mathcal{H}}$ as the space of classifiers. 

\vspace{2pt}
\noindent
\textbf{Definition 2.} Let $P_{{\bf Z}^{1:K}_i}$ denote the distribution of the unified representation that hierarchically integrates the feature vectors extracted by all feature layers. Based on $P_{{\bf Z}^{1:K}_i}$, the risk of the model can be defined by~:
\begin{equation}
\begin{aligned}
{\mathcal{E}}(P_{{\bf Z}^{1:K}_i}, h, h^\star) = {\mathbb{E}_{ {\bf z} \sim P_{{\bf Z}^{1:K}_i}}}[ {\mathcal{L}}'( h^\star( {\bf z} ), h({\bf z}) ) ]\,, 
\end{aligned}
\end{equation}
\noindent where ${\mathcal{L}}'(\cdot,\cdot)$ denotes a loss function that is symmetric and obeys the triangle inequality. $h^\star({\bf z})$ denotes a function that always returns the true class label for a given ${\bf z}$.

\vspace{2pt}
\noindent
\textbf{Theorem 1.} Let us define $h \in {\mathcal{H}}$ as a linear classifier. Let $P_{{\bf Z}^{1:K}_i}$ and $P_{{\bf Z}^{1:K}_{i+1}}$ denote two feature distributions formed using the teacher and student model, respectively. The loss function ${\mathcal{L}}'$ is bounded ${\mathcal{L}}'({\bf y}', {\bf y}') \le M$, where $({\bf y}, {\bf y}') \in {\mathcal{Y}}^2$ and $M>0$ is a positive value. Let $h^{\star}_{{\bf Z}^{1:K}_i} = \arg\min_{h \in {\mathcal{H}}}{\mathcal{E}}(P_{{\bf Z}^{1:K}_i}, h, h^\star)$ and $h^{\star}_{{\bf Z}^{1:K}_{i+1}} = \arg\min_{h \in {\mathcal{H}}}{\mathcal{E}}(P_{{\bf Z}^{1:K}_{i+1}}, h, h^\star)$ denote the ideal classifiers for the feature distributions $P_{{\bf Z}^{1:K}_{i}}$ and $P_{{\bf Z}^{1:K}_{i+1}}$, respectively. We can derive a risk bound for the model at the $(i+1)$-th task learning, expressed as~:
\begin{equation}
\begin{aligned}
{\mathcal{E}}(P_{{\bf Z}^{1:K}_i}, h, h^\star) \le {\mathcal{E}}(P_{{\bf Z}^{1:K}_{i+1}}, h, h^\star_{{\bf Z}^{1:K}_{i+1}}) + {\mathcal{D}}_{\rm disc}(P_{{\bf Z}^{1:K}_i}, P_{{\bf Z}^{1:K}_{i+1}} ) + \epsilon 
\label{eq:theorem1eq}
\,,
\end{aligned}
\end{equation}
\noindent where $\epsilon = {\mathcal{E}}(P_{{\bf Z}^{1:K}_i}, h^\star, h^\star_{{\bf Z}^{1:K}_i}) + {\mathcal{E}}(P_{{\bf Z}^{1:K}_{i+1}}, h^\star_{{\bf Z}^{1:K}_{i+1}}, h^\star_{{\bf Z}^{1:K}_i})$. ${\mathcal{D}}_{\rm disc}(P_{{\bf Z}^{1:K}_i}, P_{{\bf Z}^{1:K}_{i+1}} )$ is the discrepancy distance defined as~:
\begin{equation}
\begin{aligned}
\mathcal{D}_{\rm disc}(P_{{\bf Z}^{1:K}_i}, P_{{\bf Z}^{1:K}_{i+1}}) &= \sup_{(h,h') \in {\mathcal H}^2} \big| \mathbb{E}_{{\bf z} \sim P_{{\bf Z}^{1:K}_i} } [\mathcal{L}'(h'({\bf z}), h({\bf x}))] \\&-  \mathbb{E}_{ {\bf z} \sim P_{{\bf Z}^{1:K}_{i+1}} } [ \mathcal{L}'( h'({\bf z}), h({\bf x}) ) ] \big| \,. 
\end{aligned}
\end{equation}
The proof is provided in \cite{DomainAdaotionTheory}. In contrast to the risk bound described in \cite{DomainAdaotionTheory}, which solely assesses model performance on static target data distributions following training on static source data distributions, our proposed risk bound conceptualizes the preceding and current tasks as the source and target data distributions, respectively. This formulation enables a comprehensive analysis of forgetting phenomena within the context of continual learning. Specifically, Eq~\eqref{eq:theorem1eq} quantifies the classifier’s performance on the preceding task (${T}_i$) following the update of the model parameters from ${\theta}_i$ to ${\theta}_{i+1}$ for the new task. According to Theorem 1, a substantial discrepancy term, ${\mathcal{D}}_{\rm disc}(P_{{\bf Z}^{1:K}_i}, P_{{\bf Z}^{1:K}_{i+1}} )$, elevates the left-hand side of Eq.~\eqref{eq:theorem1eq}, which represents the target model risk. This elevation signifies a decline in model stability. To mitigate this, we introduce a novel MLFMM designed to minimize the probabilistic divergence between the feature distributions of the previous and current tasks. This approach effectively reduces the discrepancy term ${\mathcal{D}}_{\rm disc}(P_{{\bf Z}^{1:K}_i}, P_{{\bf Z}^{1:K}_{i+1}} )$, thereby enhancing model stability. Nevertheless, excessive minimization of this term may constrain the model’s adaptability to novel tasks, impairing plasticity. To address this trade-off, we propose an AROT approach that selectively penalizes alterations in critical representation layers, thereby achieving a balance between plasticity and stability.

\section{Experiments}

\subsection{Experiment Setup}

\noindent
\textbf{Task.} We evaluate the effectiveness of our proposed method on two principal scenarios: Task Incremental Learning (Task-IL) and Class Incremental Learning (Class-IL) \cite{2019Three}. In the Task-IL scenario, each training task operates with an independent label space. This means that the model is trained on distinct classes for each task, allowing it to focus solely on the features relevant to the current task's label set. During evaluation, the model receives the label space corresponding to the current task, enabling it to leverage its task-specific knowledge for accurate predictions. Conversely, in the Class-IL  scenario, the tasks share a common label space. In this setup, the model must learn to classify samples from multiple tasks simultaneously, leading to a more complex learning environment. During evaluation, the model remains unaware of the specific task to which the sample belongs; instead, it must classify the sample based on the shared label space. This requires the model to generalize its learned knowledge across tasks effectively, as it cannot rely on task-specific information during inference.

We performed our experiments across multiple datasets, specifically CIFAR-10 \cite{CIFAR10} (10 classes), CIFAR-100 (100 classes), and Tiny-ImageNet (200 classes) \cite{tinyimg}. For the CIFAR-10 dataset, we segmented it into five distinct tasks, with each task consisting of two classes. In the case of the CIFAR-100 dataset, we organized it into ten tasks, each encompassing ten classes. Additionally, for Tiny-ImageNet, we structured it into ten tasks, with each task containing twenty classes.

\noindent
\textbf{Implementation Details.} We employ ResNet18 as our foundational architecture, comprising five layers dedicated to feature extraction. To derive the adaptive regularization optimization term, we initialized five adaptive weights randomly, sourced from a uniform distribution. Our methodology extends the code from Refresh Learning \cite{refresh}, integrating our approach within their established framework. We utilized the hyperparameters specified in their implementation and conducted a grid search to ascertain the optimal value for the $\gamma$ hyperparameter, aiming to attain the most favourable outcomes.

\noindent
\textbf{Baseline.} We conducted a comparative analysis of our proposed methodology against several contemporary baselines, which encompass regularization-based techniques such as oEWC \cite{oEWC}, Synaptic Intelligence (SI) \cite{si}, Learning without Forgetting (LWF) \cite{LWF}, Classifier-Projection Regularization (CPR) \cite{CPR}, and Gradient Projection Memory (GPM) \cite{GPM}. Additionally, we incorporated Bayesian approaches like Natural Continual Learning (NCL) \cite{NCL}, architecture methods such as HAT \cite{HAT}, and memory-driven strategies including ER \cite{ER}, A-GEM \cite{A-GEM}, GSS \cite{GSS}, DER++ \cite{DER}, and HAL \cite{HAL}. Moreover, we also evaluated the most recent Refresh Learning \cite{refresh} paradigm within our comparative framework.

\subsection{Comparison of Results}
All experiments were conducted with a memory capacity of 500. For each dataset, we executed the experiments 10 times and computed the mean accuracy along with the standard deviation. The overall accuracy for both Task-IL and Class-IL scenarios is summarized in \tablename~\ref{tab:acc}.

\begin{table}[ht]
\centering
\fontsize{9}{11}\selectfont
\renewcommand{\arraystretch}{1.1} 
\caption{The average accuracy of various models on 3 datasets with buffer size 500.}
\label{tab:acc}
\vspace{5pt}
\resizebox{\textwidth}{!}{
\begin{tabular}{@{}lcccccc@{}}
    \toprule
    \multirow{2}{*}{Method} & \multicolumn{2}{c}{CIFAR-10} & \multicolumn{2}{c}{CIFAR-100} & \multicolumn{2}{c}{Tiny-ImageNet} \\
    \cmidrule(l){2-3} \cmidrule(l){4-5} \cmidrule(l){6-7} & Class-IL & Task-IL & Class-IL & Task-IL & Class-IL & Task-IL \\ \midrule 
    fine-tuning     & 19.62±0.05 & 61.02±3.33 & 09.29±0.33 & 33.78±0.42 & 07.92±0.26 & 18.31±0.68 \\
    Joint train     & 92.20±0.15 & 98.31±0.12 & 71.32±0.21 & 91.31±0.17 & 59.99±0.19 & 82.04±0.10 \\
    SI              & 19.48±0.17 & 68.05±5.91 & 09.41±0.24 & 31.08±1.65 & 06.58±0.31 & 36.32±0.13 \\
    CPR(EWC)        & 19.61±3.67 & 65.23±3.87 & 08.42±0.37 & 21.43±2.57 & 07.67±0.23 & 15.58±0.91 \\
    LWF             & 19.61±0.05 & 63.29±2.35 & 09.70±0.23 & 28.07±1.96 & 08.46±0.22 & 15.85±0.58 \\
    GPM             & -          & 90.68±3.29 & -          & 72.48±0.40 & -          & -          \\
    oEWC            & 19.49±0.12 & 64.31±4.31 & 08.24±0.21 & 21.20±2.08 & 07.42±0.31 & 15.19±0.82 \\
    NCL             & 19.53±0.32 & 64.49±4.06 & 08.12±0.28 & 20.92±2.32 & 07.56±0.36 & 16.29±0.87 \\
    HAT             & -          & 92.56±0.78 & -          & 72.06±0.50 & -          & -          \\
    UCB             & -          & 79.28±1.87 & -          & 57.15±1.67 & -          & -          \\
    HAL             & 41.79±4.46 & 84.54±2.36 & 09.05±2.76 & 42.94±1.80 & -          & -          \\
    A-GEM           & 22.67±0.57 & 89.48±1.45 & 09.30±0.32 & 48.06±0.57 & 08.06±0.04 & 25.33±0.49 \\
    GSS             & 49.73±4.78 & 91.02±1.57 & 13.60±2.98 & 57.50±1.93 & -          & -          \\
    ER              & 57.74±0.27 & 93.61±0.27 & 20.98±0.35 & 73.37±0.43 & 09.99±0.29 & 48.64±0.46 \\
    DER++           & 72.70±1.36 & 93.88±0.50 & 36.37±0.85 & 75.64±0.60 & 18.90±0.09 & 51.84±0.47 \\
    DER++ + Refresh & 73.88±1.16 & 94.44±0.39 & 39.10±0.65 & 76.80±0.31 & 16.16±0.72 & 53.36±1.25 \\
    OWMMD (Ours)    & \textbf{75.29±0.75} &\textbf{94.94±0.41} & \textbf{41.75±0.58} & \textbf{76.99±0.38} & \textbf{19.18±0.31} & \textbf{53.69±0.93} \\ \bottomrule
\end{tabular}}
\end{table}

In comparison to fine-tuning and Joint train, OWMMD excels because these baselines are unable to effectively address catastrophic forgetting when exposed to sequential tasks. Fine-tuning shows poor performance due to its inability to retain past knowledge when exposed to new tasks, resulting in a steady decline in accuracy as more tasks are introduced. Joint train, while providing a notable improvement, still struggles in Class-IL settings, where it does not exhibit the task-specific adaptation necessary for continual learning. In contrast, OWMMD integrates not only MLFMM and AROT but also memory mechanisms, which allow the model to not only adapt to new tasks but also retain essential knowledge from previously learned tasks.

While regularization-based methods such as SI, LWF, and CPR improve performance over fine-tuning, they still fall short in preventing forgetting over a long sequence of tasks. These methods attempt to regularize the model parameters, but their capacity to effectively manage the trade-off between learning new tasks and retaining old ones is limited. For instance, while SI and CPR perform reasonably well in Task-IL settings, they still underperform in Class-IL scenarios due to their inability to fully accommodate the growing complexity of continual task learning. On the other hand, OWMMD surpasses these methods by utilizing an adaptive regularization term, ensuring that each model layer contributes appropriately to both new and old tasks, thus resulting in improved performance across all datasets, especially in Class-IL settings.

Memory-based methods such as ER and A-GEM are designed to mitigate forgetting by storing and reusing past experiences. While ER performs well in Task-IL, where it can replay past data to avoid forgetting, and A-GEM constrains updates to avoid drastic shifts in learned knowledge, both methods still face limitations in handling large-scale, complex datasets. OWMMD outperforms both ER and A-GEM, particularly in Task-IL, due to its ability to dynamically adapt the regularization weights during training. This ensures that the model does not overfit to old tasks while still retaining key knowledge when presented with new tasks.

When comparing OWMMD to more advanced methods like DER++, Refresh Learning, and HAT, our approach consistently delivers superior results, particularly on CIFAR-10 and CIFAR-100 in the Class-IL setting. While DER++ and Refresh Learning both exhibit competitive performance, OWMMD surpasses them due to its more robust handling of task-specific regularization and memory management. DER++ shows a strong performance across all datasets, but OWMMD outperforms it on CIFAR-10 and CIFAR-100, where our adaptive regularization and memory strategy provide more effective balancing of knowledge retention. Refresh Learning performs well, but it does not match the performance of OWMMD on these datasets, particularly in the Task-IL scenario, where our method achieves a higher accuracy.

In summary, OWMMD demonstrates its superiority over a wide range of baselines due to its innovative integration of adaptive regularization and memory mechanisms. These features allow our model to effectively mitigate catastrophic forgetting, retain crucial knowledge, and adapt to new tasks, outperforming both classical methods and more recent continual learning frameworks across multiple datasets and settings.

\noindent
\textbf{Backward Transfer Analysis.}
Backward Transfer (BWT) measures the effect of learning new tasks. Negative values indicate a decline in performance on earlier tasks after training on new tasks, highlighting the challenge of catastrophic forgetting in continual learning. \tablename~\ref{tab:bwt} presents the BWT scores of various methods across three datasets: CIFAR-10, CIFAR-100 and Tiny-ImageNet, in both Class-IL and Task-IL scenarios. 

In the case of fine-tuning, the BWT values are substantially negative across all datasets, with CIFAR-10 and Tiny-ImageNet exhibiting values as low as -96.39 and -78.94, respectively. This suggests that fine-tuning on new tasks causes significant forgetting of previously learned tasks. Similarly, HAL and A-GEM, while better than fine-tuning, still exhibit considerable negative BWT, especially in Class-IL settings, with HAL showing -62.21 for CIFAR-10 and A-GEM showing -94.01 for CIFAR-10. These values indicate that these methods fail to effectively mitigate forgetting when new tasks are introduced, resulting in a performance drop on earlier tasks.

In contrast, OWMMD demonstrates remarkably lower BWT values, with the best performance seen in Task-IL settings, where the BWT is -3.0 on CIFAR-10. The method significantly reduces backward transfer on both CIFAR-100 and Tiny-ImageNet, with BWT values of -42.99 and -59.31, respectively, compared to higher values in other methods. This indicates that the adaptive regularization mechanism of OWMMD helps minimize the degradation of previously learned knowledge, allowing the model to retain crucial information even after the introduction of new tasks.

Other methods like ER and DER++ also show some improvement in backward transfer, but they still suffer from a noticeable decline in performance on earlier tasks. Refresh Learning demonstrates strong results, especially on CIFAR-10 and CIFAR-100, but OWMMD still outperforms it with lower BWT values, highlighting the advantages of our proposed adaptive regularization strategy in mitigating forgetting.

\begin{table}[ht]
\centering
\fontsize{9}{11}\selectfont
\renewcommand{\arraystretch}{1.1} 
\caption{Backward Transfer of various methods on 3 datasets with buffer size 500.}
\label{tab:bwt}
\vspace{5pt}
\resizebox{\textwidth}{!}{
\begin{tabular}{@{}lcccccc@{}}
    \toprule
    \multirow{2}{*}{Method} & \multicolumn{2}{c}{CIFAR-10} & \multicolumn{2}{c}{CIFAR-100} & \multicolumn{2}{c}{Tiny-ImageNet} \\
    \cmidrule(l){2-3} \cmidrule(l){4-5} \cmidrule(l){6-7} & Class-IL & Task-IL & Class-IL & Task-IL & Class-IL & Task-IL \\ \midrule 
    fine-tuning & -96.39±0.12 & -46.24±2.12 & -89.68±0.96 & -62.46±0.78 & -78.94±0.81 & -67.34±0.79 \\
    HAL        & -62.21±4.34 & -05.41±1.10 & -49.29±2.82 & -13.60±1.04 & - & - \\
    A-GEM      & -94.01±1.16 & -14.26±1.18 & -88.50±1.56 & -45.43±2.32 & -78.03±0.78 & -59.28±1.08 \\
    GSS        & -62.88±2.67 & -07.73±3.99 & -82.17±4.16 & -33.98±1.54 & - & - \\
    ER         & -45.35±0.07 & -03.54±0.35 & -74.84±1.38 & -16.81±0.97 & -75.24±0.76 & -31.98±1.35 \\
    DER++      & -22.38±4.41 & -04.66±1.15 & -53.89±1.85 & -14.72±0.96 & -60.71±0.69 & -29.01±0.46 \\
    DER++~+~Refresh    & -22.03±3.89 & -04.37±1.25 & -53.51±0.70 & -14.23±0.75 & -65.07±0.73 & \textbf{-27.36±1.49} \\
    OWMMD(ours)     & \textbf{-20.61±0.45} & \textbf{-03.00±0.48} & \textbf{-42.99±1.23} & \textbf{-13.80±0.84}  & \textbf{-59.31±0.64} & -27.73±1.00  \\ \bottomrule
\end{tabular}}
\end{table}

Overall, the results emphasize that OWMMD stands out as the most effective approach in terms of minimizing backward transfer, addressing the challenge of catastrophic forgetting, and ensuring that the model can learn new tasks without significant degradation in the performance of previously learned tasks.

\subsection{Comparison of Results for Complex Datasets}
In the comparison on complex datasets, as shown in \tablename~\ref{tab:acc2}, OWMMD demonstrates superior performance across all three datasets (Cars-196, Cub-200, and ImageNet-R), significantly outperforming other methods. On the Cars-196 and Cub-200 datasets, both in Class-IL and Task-IL settings, OWMMD leads by a large margin, particularly in the Task-IL setting, with accuracies of 52.53\% and 48.69\% respectively, far surpassing other methods like DER++ (36.59\% and 37.68\%) and ER (26.06\% and 28.71\%). On the ImageNet-R dataset, OWMMD similarly shows strong robustness, with accuracies of 8.93\% in Class-IL and 30.39\% in Task-IL, outperforming DER++ (5.82\% and 22.73\%) and DER (3.88\% and 17.17\%) among others.

\begin{table}[ht]
\centering
\fontsize{9}{11}\selectfont
\renewcommand{\arraystretch}{1.1} 
\caption{Average accuracy of various models on 3 complex datasets with buffer size 500.}
\vspace{5pt}
\resizebox{\textwidth}{!}{
\begin{tabular}{@{}lcccccc@{}}
    \toprule
    \multirow{2}{*}{Method} & \multicolumn{2}{c}{Cars-196} & \multicolumn{2}{c}{Cub-200} & \multicolumn{2}{c}{ImageNet-R} \\
    \cmidrule(l){2-3} \cmidrule(l){4-5} \cmidrule(l){6-7} & Class-IL & Task-IL & Class-IL & Task-IL & Class-IL & Task-IL \\ \midrule 
    LWF             & 07.27±0.13 & 11.99±0.05 & 04.89±0.10 & 09.95±0.49 & 04.93±0.19 & 11.55±0.39 \\
    oEWC            & 05.22±0.62 & 10.53±0.38 & 04.34±0.12 & 09.83±0.45 & 04.28±0.52 & 12.98±0.82 \\
    ER              & 05.63±0.09 & 26.06±1.19 & 07.05±0.34 & 28.71±0.63 & 03.94±0.82 & 19.68±3.97 \\
    DER             & 07.92±0.70 & 31.50±0.34 & 06.79±0.37 & 29.87±0.86 & 03.88±0.76 & 17.17±3.31 \\
    DER++           & 09.34±0.33 & 36.59±1.19 & 13.50±3.11 & 37.68±3.89 & 05.82±0.64 & 22.73±2.23 \\
    DER++ + Refresh & 08.70±0.57 & 35.63±1.00 & 13.59±0.74 & 37.79±0.66 & 05.47±0.49 & 22.50±1.12 \\
    OWMMD (Ours)    & \textbf{18.09±0.86} & \textbf{52.53±1.24} & \textbf{22.84±2.49} & \textbf{48.69±1.37} & \textbf{08.93±1.14} & \textbf{30.39±2.33} \\ \bottomrule
\end{tabular}}
\label{tab:acc2}
\end{table}

These results indicate that OWMMD not only achieves excellent results on standard datasets but also excels when faced with more challenging and complex tasks. Especially in preventing catastrophic forgetting and maintaining high accuracy, OWMMD stands out. Compared to other methods, OWMMD is particularly effective in multi-task settings, adapting better to the variability and complexity of complex datasets, offering a more efficient and robust solution for continual learning tasks.

\subsection{Compared to the Pre-Trained Models (PTMs)}

Recent investigations have examined the pre-trained Vision Transformer (ViT) \cite{ViT} as the core backbone architecture, incorporating a dynamically generated, lightweight expert module for learning new tasks. These pre-trained model (PTM)-based methodologies have demonstrated superior effectiveness in continual learning scenarios by leveraging the rich representations derived from PTMs. In this section, we extend the proposed OWMMD approach to PTM-based architectures with the objective of augmenting predictive accuracy and assessing the scalability of OWMMD. Specifically, consistent with \cite{DECM}, we utilize a pre-trained ViT as the foundational backbone and design an expert module comprising a feature extractor and a linear classifier. Upon encountering a novel task, a new expert module is instantiated dynamically, enabling focused learning of the task at hand, while all previously learned experts remain frozen to prevent catastrophic forgetting. To mitigate computational complexity, only the last three feature layers of the backbone are optimized during training. After completing each task, the primary backbone is duplicated and set to a frozen state, functioning as a teacher model. The OWMMD method is then applied to regulate the optimization process of the backbone's parameters using the teacher module and the current active backbone. This modified approach, designated as OWMMD-PTMs, is evaluated against recent dynamic expansion models employing pre-trained ViT architectures, with detailed results presented in Table~\ref{tab:standardAcc}. Empirical findings indicate that OWMMD-PTMs consistently surpasses baseline methods across various datasets, substantiating its capacity to enhance model performance through the integration with pre-training strategies.

\begin{table}[t]
    \centering
      \caption{The average accuracy calculated by various models on standard continual learning benchmarks, as averages of 10 runs. The results of baselines are taken from \cite{CL_DarkKD} and \cite{UnifiedCL,InteractiveCL, DECM}.}
      \vspace{5pt}
\fontsize{9}{11}\selectfont\setlength{\tabcolsep}{0.75mm}{
\begin{tabular}{@{}l c c  @{}}
\toprule 
 \textBF{Methods}& TinyImageNet& CIFAR100
\\
\midrule 
ICL w Pure-MM \cite{InteractiveCL}&-&96.35
\\
KAMO \cite{DECM} &92.05 $\pm$ 0.42 &97.86 $\pm$ 0.37 \\
DualPrompt \cite{DualPrompt} &-&93.76
\\
L2P \cite{LearningPrompt} &-&93.92
\\
OWMMD-PTMs&\textBF{94.15} $\pm$ 0.38&\textBF{98.24} $\pm$ 0.32
\\
\bottomrule 
 \end{tabular}
\label{tab:standardAcc}
 }
\end{table}

\subsection{Analysis Results}

\begin{figure}[htp]
\centering
\setlength{\abovecaptionskip}{0.05cm}
\begin{minipage}[b]{\textwidth}
  \centering
  \includegraphics[width=\textwidth]{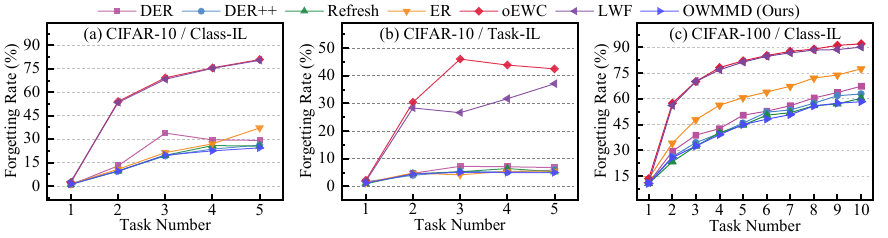}
\end{minipage}
\vfill
\begin{minipage}[b]{\textwidth}
  \centering
  \includegraphics[width=\textwidth]{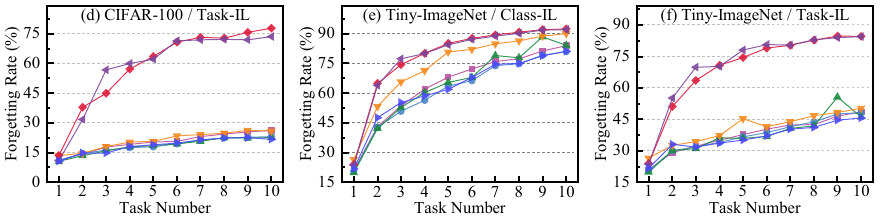}
\end{minipage}
\caption{Forgetting curve analysis of various models on CIFAR-10 , CIFAR-100 and Tiny-ImageNet datasets under class-incremental and task-incremental learning settings. }
\label{fig:forgetting_curves}
\end{figure}

\noindent
\textbf{Forgetting Curve.}
Forgetting rates highlights the performance and knolwedge losses over time. In this experiment, we investigate the forgetting curves for several methods on the CIFAR-10, CIFAR-100 and Tiny-ImageNet datasets under both class-incremental learning (Class-IL) and task-incremental learning (Task-IL), and the results are presented in \figurename~\ref{fig:forgetting_curves}. The empirical results demonstrate that different methods show different forgetting behaviors.

In CIFAR-10, OWMMD outperforms all other methods under both Class-IL and Task-IL settings, showing a notably lower forgetting rate. This is particularly evident in the task progression where OWMMD maintains stable performance over time, while other methods, such as DER and ER, experience more significant forgetting. A similar trend holds for CIFAR-100 and Tiny-ImageNet, where OWMMD exhibits superior stability across tasks, demonstrating its effectiveness in retaining knowledge from earlier tasks. Furthermore, the proposed OWMMD achieves the lowest forgetting rate than other baselines at each task learning, demonstrating its strong ability to fight forgetfulness. 

\noindent
\textbf{Optimal Weight Analysis.}
In the proposed adaptive regularization optimization term, as described in Section \ref{sec:aw}, we introduce adaptive weights for each network layer. These weights are dynamically adjusted to optimize knowledge transfer between the teacher and student networks. During each task learning, the model can determine the optimal weights specific to the current task. These weights are updated and refined as the network progresses through subsequent tasks, ensuring continuous improvement in the learning process.

\begin{figure}[htp]
    \centering
    \setlength{\abovecaptionskip}{0.05cm}
    \includegraphics[width=\textwidth]{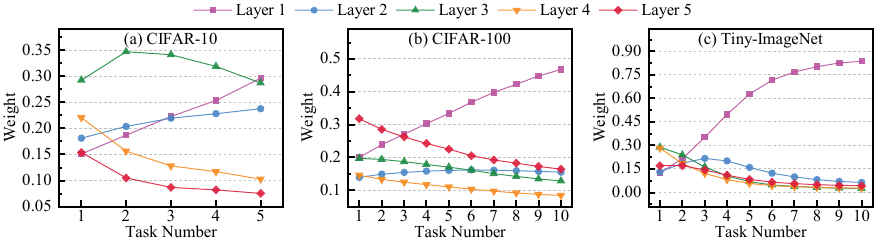}
     \caption{Dynamic Adjustment Process of Layer-wise Adaptive Weights Across Tasks. }
    \label{fig:weights1}
\end{figure}

The evolution of these weights across tasks is illustrated in \figurename~\ref{fig:weights1}. From these results, we can find several trends, summarized in the following~:
\begin{enumerate}
    \item \textbf{Shallow Layers:} The weights assigned to the shallower layers tend to increase as the tasks progress. This increase suggests that the network relies more heavily on basic, lower-level features as new tasks are introduced. These shallow layers capture fundamental representations, such as edges and textures, which are generally useful across different tasks, providing semantically rich information that supports continual learning.
    \item \textbf{Deep Layers:} The weights for deeper layers show a decreasing trend across both datasets. This result indicates that as more tasks are added, the network becomes less dependent on features extracted by these deeper layers. This reflects the need to retain generalized knowledge that is applicable across a range of tasks rather than retaining highly specialized features, which may be less transferable.
    \item \textbf{Mid-Level Layers:} The mid-level layers demonstrate an initial increase in weight, followed by a decline as training progresses. For instance, Layer 3 in CIFAR-10 and Layer 2 in Tiny-ImageNet both initially gain importance as the network learns to handle the variety of tasks introduced early on. This increase occurs because these layers capture moderately complex features that are valuable during the initial stages of learning. However, as more tasks are introduced, these mid-level features become less significant compared to the foundational representations provided by the shallow layers. The eventual decline in mid-level layer importance reflects the network's shift towards retaining the most stable and general features over task-specific knowledge.
\end{enumerate}

\begin{figure}[htb]
    \centering
    \setlength{\abovecaptionskip}{0.05cm}
    \includegraphics{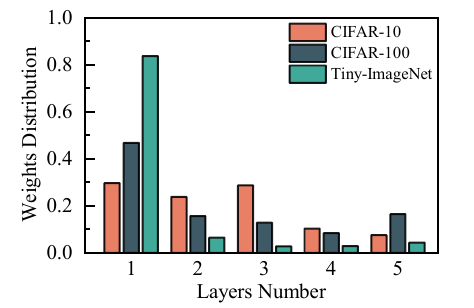}
    \caption{The distribution of the final adaptive weights after learning all tasks.}
    \label{fig:weights2}
\end{figure}

\figurename~\ref{fig:weights2} illustrates the final distribution of adaptive weights between the different layers of the network after completing all tasks. It is evident that the optimal weight distribution varies significantly between datasets such as CIFAR-10 and Tiny-ImageNet. For CIFAR-10, the network tends to assign higher weights to certain shallow layers, indicating their importance in generalizing across tasks. On the other hand, for Tiny-ImageNet, the weight distribution is more concentrated towards specific layers, reflecting the dataset's complexity and the unique feature extraction requirements. These differences highlight how the network adapts its internal representations depending on the characteristics of the dataset, dynamically adjusting the importance of each layer to optimize knowledge retention and transfer in a continual learning setting.

\noindent
\textbf{Distance Function Analysis.}
\tablename~\ref{tab:distance} compares the performance of different distance functions and the effect of using optimal weights in our experiments on CIFAR-10 with a buffer size of 500. The first column represents the distance function used, while the second column indicates whether the adaptive regularization optimization term (AROT) was applied, with "False" denoting no use of adaptive weights and "True" indicating their use.

From the results, it is evident that using the Maximum Mean Discrepancy (MMD) distance function yields the best performance in both the Class-IL and Task-IL scenarios. Specifically, the configuration with MMD and adaptive weights achieves the highest accuracy. Additionally, we observe that the use of adaptive weights consistently outperforms the fixed weight configurations. This demonstrates the effectiveness of both MMD as a distance function and the benefits of applying adaptive weights to improve performance during continual learning.

\begin{table}[htp]
    \centering
    \fontsize{9}{11}\selectfont
    \renewcommand{\arraystretch}{1.1} 
    \caption{Comparison of different distance functions and the effect of adaptive regularization optimization term (AROT) on CIFAR-10 with buffer size 500. The first column shows the distance used (cos, L2, or MMD). The second column indicates whether the adaptive weights (AROT) were applied, where "True" denotes the use of adaptive weights and "False" means no adaptive weights were applied. The last two columns display the performance results for Class-IL and Task-IL scenarios, including mean accuracy with standard deviation. The highest performance in each case is marked in bold.}
    \vspace{5pt}
    \begin{tabular}{@{}cccc@{}} 
    \toprule
    Distance & AROT  & Class-IL   & Task-IL    \\ 
    \midrule
    Cosine   & False & 74.62±0.79 & 94.53±0.18 \\
    Cosine   & True  & 74.89±0.86 & 94.67±0.27 \\
    L2       & False & 75.12±0.74 & 94.88±0.33 \\
    L2       & True  & 75.23±0.94 & 94.56±0.15 \\
    MMD      & False & 74.69±1.07 & 94.51±0.24 \\
    MMD      & True  & \textbf{75.29±0.75} & \textbf{94.94±0.41} \\ 
    \bottomrule
    \end{tabular}
    \label{tab:distance}
\end{table}

\noindent
\textbf{Buffer Size.}
Table~\ref{tab:acc_buffer} shows the performance of our proposed OWMMD method with different buffer sizes compared with continual learning baselines. Evaluations on CIFAR-10 under Class-IL and Task-IL settings reveal three key observations:

With the smaller buffer size (200), OWMMD achieves the highest accuracy in both scenarios (67.74\% Class-IL, 92.97\% Task-IL), outperforming all compared methods. Notably, its Class-IL accuracy surpasses the second-best DER+++Refresh by 2.35\%, while maintaining a 0.17\% Task-IL advantage. This demonstrates superior capability in memory-constrained environments where methods like GEM (25.54\% Class-IL) and A-GEM (20.04\% Class-IL) show significant degradation.

When using the 5120 buffer, OWMMD maintains competitiveness with 86.33\% Class-IL accuracy (best overall) and 96.61\% Task-IL accuracy. Although ER achieves marginally higher Task-IL performance (96.98\%), it suffers a 14.51\% accuracy drop in Class-IL compared to its Task-IL results, while OWMMD shows balanced performance with only 10.3\% difference.

The progressive improvement from 200 to 5120 buffers confirms OWMMD's effective memory utilization: Class-IL accuracy increases by 18.59\% absolute (67.74\% $\rightarrow$ 86.33\%), significantly exceeding the average 15.2\% gain of other methods. This demonstrates our method's unique advantage in scenarios requiring flexible memory scaling without performance saturation.
\begin{table}[ht]
\centering
\fontsize{9}{11}\selectfont
\renewcommand{\arraystretch}{1.1} 
\caption{Performance comparison with buffer size 500 as baseline. Arrows (↑/↓) indicate performance changes relative to buffer size 500. Our method shows progressive improvement with larger buffers.}
\label{tab:acc_buffer}
\vspace{5pt}
\resizebox{\textwidth}{!}{
\begin{tabular}{@{}lcccccc@{}}
    \toprule
    \multirow{2}{*}{Method} & \multicolumn{2}{c}{200} & \multicolumn{2}{c}{500} & \multicolumn{2}{c}{5120} \\ 
    \cmidrule(lr){2-3} \cmidrule(lr){4-5} \cmidrule(lr){6-7}
    & Class-IL & Task-IL & Class-IL & Task-IL & Class-IL & Task-IL \\ 
    \midrule 
    GEM         & 25.54±0.76↓ & 90.44±0.94↓ & 26.20±1.26 & 92.16±0.69 & 25.26±3.46↓ & 95.55±0.02↑ \\
    iCaRL       & 49.02±3.20↑ & 88.99±2.13↑ & 47.55±3.95 & 88.22±2.62 & 55.07±1.55↑ & 92.23±0.84↑ \\
    FDR         & 30.91±2.74↑ & 91.01±0.68↓ & 28.71±3.23 & 93.29±0.59 & 19.70±0.07↓ & 94.32±0.97↑ \\
    HAL         & 32.36±2.70↓ & 82.51±3.20↓ & 41.79±4.46 & 84.54±2.36 & 59.12±4.41↑ & 88.51±3.32↑ \\
    A-GEM       & 20.04±0.34↓ & 83.88±1.49↓ & 22.67±0.57 & 89.48±1.45 & 21.99±2.29↓ & 90.10±2.09↑ \\
    GSS         & 39.07±5.59↓ & 88.80±2.89↓ & 49.73±4.78 & 91.02±1.57 & 67.27±4.27↑ & 94.19±1.15↑ \\
    ER          & 44.79±1.86↓ & 91.19±0.94↓ & 57.74±0.27 & 93.61±0.27 & 82.47±0.52↑ & \textbf{96.98±0.17↑} \\
    DER++       & 64.88±1.17↓ & 91.92±0.60↓ & 72.70±1.36 & 93.88±0.50 & 85.24±0.49↑ & 96.12±0.21↑ \\
    DER+++Refresh & 65.39±1.01↓ & 92.80±0.42↓ & 73.88±1.16 & 94.44±0.39 & 85.98±0.43↑ & 96.43±0.11↑ \\
    OWMMD (Ours)& \textbf{67.74±1.06↓} & \textbf{92.97±0.37↓} & \textbf{75.29±0.75} & \textbf{94.94±0.41} & \textbf{86.33±0.37↑} & 96.61±0.13↑ \\ 
    \bottomrule
\end{tabular}}
\end{table}

\noindent
\textbf{Training Time Comparison.}
In terms of training time, OWMMD demonstrates significant efficiency improvements on CIFAR-10 and CIFAR-100 compared to many baselines. For CIFAR-10, OWMMD completes training in 143 minutes—2.4× faster than GSS (341 minutes) and nearly twice as fast as GEM (280 minutes). While lightweight methods like LWF (32 min) and oEWC (28 min) require less time, OWMMD maintains a practical training duration that remains competitive with most approaches, particularly in scenarios involving multiple sequential tasks. This efficiency advantage persists on CIFAR-100, where OWMMD (149 min) trains 2.7× faster than GSS (395 min) and 4.7× faster than GEM (705 min), while retaining strong performance.

\begin{figure}[htp]
    \centering
    \setlength{\abovecaptionskip}{0.05cm}
    \includegraphics[width=\textwidth]{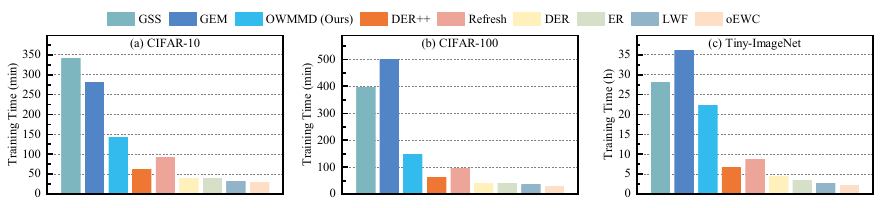}
     \caption{Training time comparison of various models on CIFAR-10/CIFAR-100 (50 epochs) and Tiny-ImageNet (100 epochs) Datasets with RTX 4090D GPU.}
    \label{fig:training_time}
\end{figure}

For Tiny-ImageNet, OWMMD requires 22.3 hours, which exceeds DER++ (6.63 hours) and LWF (2.54 hours) by 3.4× and 8.8× respectively. However, this increased computational cost is offset by OWMMD’s superior accuracy and stability in continual learning scenarios. The method’s ability to mitigate catastrophic forgetting while maintaining competitive runtime efficiency, despite the dataset’s complexity, positions it as a robust solution for continual learning tasks.

\subsection{Ablation Study}
To analyze the effectiveness of multi-layer feature matching and adaptive regularization, we conduct an ablation study by selectively applying MMD to different layers of the backbone. Specifically, we compare models that use only a subset of layers (e.g., the first few or last few) against the full model, and we evaluate the impact of the Adaptive Regularization Optimization Term (AROT). This experiment is performed on CIFAR-10 with a buffer size of 500.

\tablename~\ref{tab:ablation} shows that when AROT is enabled, using three layers for MMD regularization yields better performance than using only two, demonstrating that a broader feature alignment contributes to more effective continual learning. Notably, aligning the first three layers achieves superior results compared to the last three layers, indicating that shallow feature matching plays a more significant role in mitigating forgetting. This aligns with the observed adaptive weight trends, where lower-level layers receive higher importance as training progresses. Shallow layers capture fundamental structures that are transferable across tasks, whereas deeper layers extract task-specific features that may not generalize as effectively. Nevertheless, the best results are obtained when all layers are utilized with AROT, as this ensures a well-balanced integration of generalizable and task-specific representations.

When AROT is not used, performance degrades across all layer configurations, underscoring the importance of adaptive weighting. Without AROT, models exhibit greater fluctuations in accuracy depending on the selected layers, as fixed weighting fails to optimally balance the contributions of different feature levels. The consistent improvement observed with AROT highlights its role in dynamically adjusting the influence of each layer, enabling better feature retention and mitigating catastrophic forgetting.

\begin{table}[ht]
    \centering
    \fontsize{9}{11}\selectfont
    \renewcommand{\arraystretch}{1.1} 
    \caption{Ablation study on CIFAR-10 with buffer size 500. "First $n$" indicates that only the first $n$ layers of the backbone are utilized for matching and MMD calculation, while "Last $n$" refers to using only the last $n$ layers. "ALL" indicates that all backbone layers are used. AROT denotes Adaptive Regularization Optimization Term.}
    \vspace{5pt}
    \setlength{\tabcolsep}{3mm}{
        \begin{tabular}{@{}ccccc@{}}
        \toprule
        \multirow{2}{*}{Layers} & \multicolumn{2}{c}{With AROT} & \multicolumn{2}{c}{No AROT} \\ \cmidrule(lr){2-3} \cmidrule(lr){4-5} 
            & Class-IL      & Task-IL       & Class-IL     & Task-IL      \\
        \midrule
        First 1  & -             & -             & 73.94±0.96   & 94.38±0.38   \\
        First 2  & 74.57±0.57    & 94.41±0.61    & 74.04±0.59   & 94.56±0.44   \\
        First 3  & 74.77±0.99    & 94.39±0.54    & 73.25±0.72   & 94.42±0.17   \\
        Last 3   & 74.80±0.38    & 94.44±0.06    & 74.46±0.79   & 94.42±0.45   \\
        Last 2   & 74.36±0.87    & 94.50±0.29    & 74.06±0.79   & 94.38±0.06   \\
        Last 1   & -             & -             & 74.87±1.04   & 94.30±0.38   \\
        ALL & \textbf{75.29±0.75} & \textbf{94.94±0.41} & 74.69±1.07 & 94.51±0.24 \\
        \bottomrule
        \end{tabular}
    }
    \label{tab:ablation}
\end{table}

\section{Conclusion and Future Works}
In this paper, we introduce an innovative framework for alleviating catastrophic forgetting in continual learning, called  Optimally-Weighted Maximum Mean Discrepancy (OWMMD). Specifically, the proposed framework introduces a Multi-Level Feature Matching Mechanism (MLFMM) to penalize representation changes in order to relieve network forgetting. This technique utilizes various layers of neural networks to ensure coherence between the feature representations of prior and current tasks. By integrating a regularization term that reduces the disparity between feature representations across different tasks, our methodology illustrates the capability to avert catastrophic forgetting while facilitating the model's adaptation to a dynamic data stream.

The empirical findings indicate that our methodology surpasses current techniques regarding both precision and consistency across various tasks. By modulating the feature space and accommodating new tasks while preserving previously acquired knowledge, our approach enhances the expanding literature on continual learning. Furthermore, the incorporation of probabilistic distance metrics in feature alignment offers a promising avenue for advancing task generalization and transfer learning.

Task-Free Continual Learning (TFCL) presents a more complex paradigm, characterized by the absence of explicit task delineation and boundaries throughout both training and evaluation stages. Given that the proposed OWMMD framework utilizes reservoir sampling for dynamic memory buffer updates, its memory management strategy is inherently compatible with TFCL settings. Nevertheless, a notable constraint in adapting OWMMD to TFCL lies in the reliance of the MLFMM component on task-specific information to archive the current model as a teacher module for regularization loss computation. To overcome this limitation, we would like to develop a straightforward yet effective task-boundary detection technique that leverages the emergence of novel categories as indicators of task transitions. This direction will be systematically explored in our future research. We have added the discussion in the conclusion and limitation section of the revised version.

While promising, there are several areas where this work can be expanded in future research. First, the efficiency of memory buffer management and sample selection can be further optimized by exploring more advanced sampling strategies or leveraging external memory architectures. Second, our method's performance on larger-scale datasets and in real-world applications, such as robotics or autonomous driving, remains to be fully explored. Finally, extending our feature-matching approach to other types of neural architectures, such as transformer-based models or generative networks, could provide broader applicability and improve performance across a wider range of tasks.

In summary, this work lays the foundation for more robust continual learning systems that can efficiently retain and transfer knowledge across multiple tasks. Future work will focus on refining the proposed methods, exploring additional regularization techniques, and evaluating the approach in more complex settings to advance the state-of-the-art in continual learning.





\bibliographystyle{elsarticle-num} 
\bibliography{VAEGAN}






\end{document}